\documentclass[conference]{ieeeconf}  

\IEEEoverridecommandlockouts                              
\usepackage{authblk}
\usepackage{algpseudocode}
\usepackage{algorithm}
\usepackage{optidef}
\usepackage{graphics} 
\usepackage{epsfig} 
\usepackage{mathptmx} 
\usepackage{times} 
\usepackage{amsmath} 
\usepackage{amssymb}  
\usepackage{t1enc}
\usepackage[bookmarks=false,hidelinks]{hyperref}
\usepackage{tikz}
\usepackage[caption=false,font=footnotesize]{subfig}
\usetikzlibrary{decorations.pathmorphing}
\usetikzlibrary{shapes,arrows,shadows,patterns}
\usepackage[printonlyused]{acronym}
\usepackage{nomencl}
\usepackage{mathrsfs}
\newcommand{\norm}[1]{\left\lVert#1\right\rVert}

\usepackage{fancyhdr}
\usepackage{kantlipsum}

\usepackage{eso-pic}

\begin{document}



\title{\LARGE \bf
Collision-Free Multi Robot Trajectory Optimization in Unknown Environments using Decentralized Trajectory Planning}


\author{Vijay Arvindh, Govind Aadithya R, Shravan Krishnan and Sivanathan K  
\thanks{This work was supported by SRM Institute of Science and Technology}
\thanks{ The authors are with Autonomous Systems Lab, SRM Institute of Science and Technology,India} \thanks{Email:\textit{vijay\_arvindh,govinda\_adithya,shravan\_krishnan)@srmuniv.edu.in, sivanathan.k@ktr.srmuniv.ac.in}}
\thanks{\textit{Corresponding Author: Shravan Krishnan}}
}
\maketitle

\begin{abstract}
Multi robot systems have the potential to be utilized in a variety of applications. In most of the previous works, the trajectory generation for multi robot systems is implemented in known environments. To overcome that we present an online  
trajectory optimization algorithm that utilizes communication of robots' current states to account to the other robots while using local object based maps for identifying obstacles. Based upon this data, we predict the trajectory expected to be traversed by the robots and utilize that to avoid collisions by formulating regions of free space that the robot can be without colliding with other robots and obstacles. A trajectory is optimized constraining the robot to remain within this region.The proposed method is tested in simulations on Gazebo using ROS.  \\

\end{abstract}
\vspace{2mm}

\section{Introduction}
Multi-robots systems are a group of individual entities working together so as to maximize their own performance while accounting for some higher goals. The trajectories generated in such scenarios will have to ensure that the robots don't collide with one another while keeping up with their dynamic limits. The trajectory generation process in multi agent systems has long since been done in a centralized manner wherein the trajectories are generated before hand and transmitted across to individual robots. These methods show excellent performance and safety when the number of robots are known beforehand and limited in number as scalability is a problem in them. Recently, this has branched out to decentralized approaches that attempt to plan trajectories in known environments using a variety of different approaches, but it is important to mitigate disturbances and unmodeled errors by modifying the trajectories during run time, making it more like an implicit feedback.  
\begin{figure}
\includegraphics[width=0.5\textwidth]{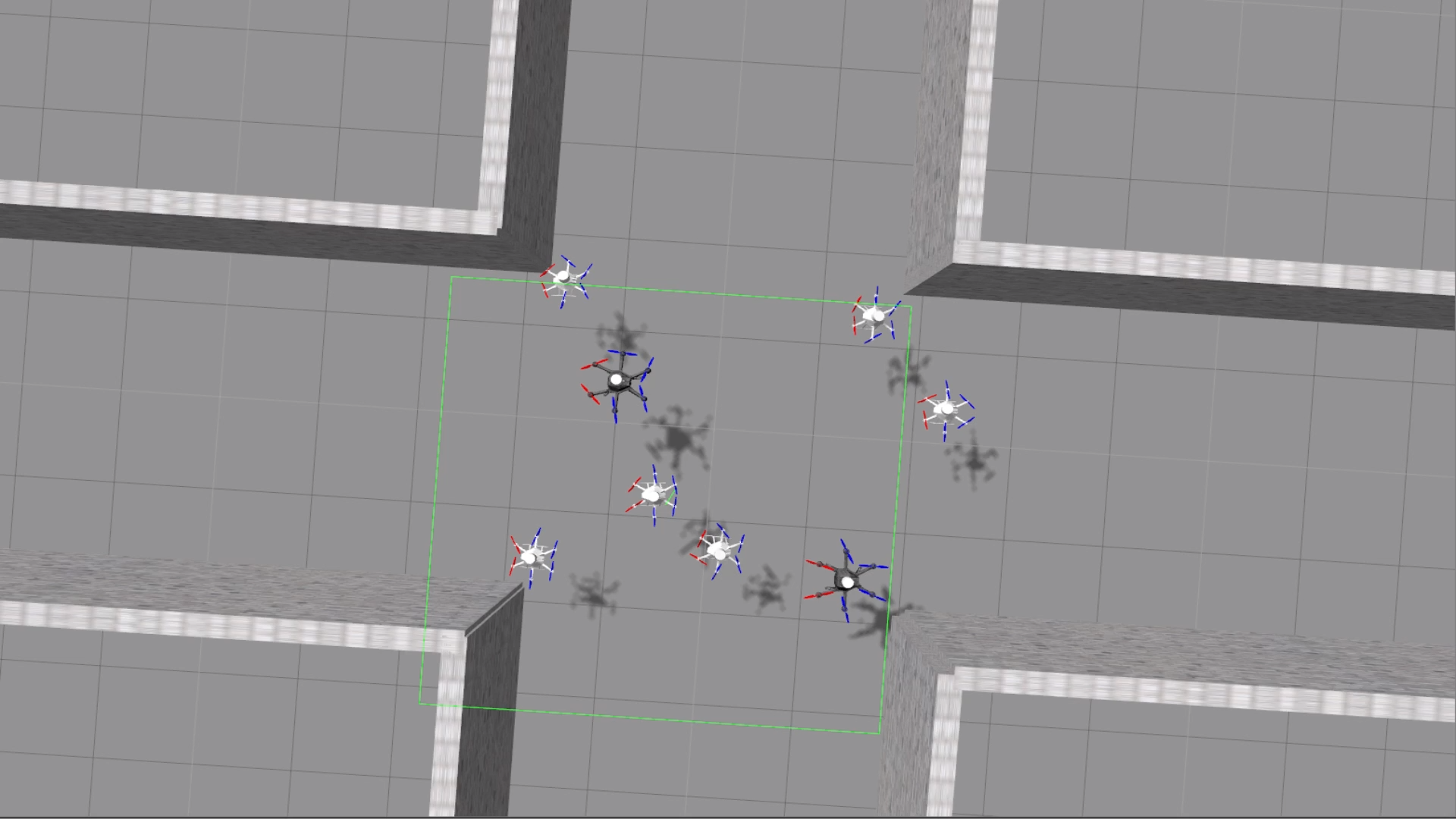}
\caption{Six AscTec Fireflys and two Neos at an Intersection-like Environment}
\label{firefly6}
\end{figure}

Our approach attempts to solve this problem using two steps, i) Collision-free regions that the robot can be safe within to avoid collisions with other robots and ii) Generate a trajectory for the robot constraining it within the safe region. Other obstacles also have to be considered while planning the trajectory. To achieve this the obstacles are stored using primitive model representations and used as soft constraint for the objective with the obstacles approximated as circles of appropriate radii. The approach exploits differential flatness \cite{fliess1995flatness} based trajectory generation for n\textsuperscript{th} order systems  property of robots \cite{mellinger2011minimum}, \cite{optimalwalambe2016}. and constrains the generated trajectory to stay within a safe region at specified discrete time points in two dimensional environments. 

The contributions can be stated as
\begin{enumerate}
\item A decentralized trajectory optimization algorithm for multi robot systems.
\item A simple method for obstacle representation in 2D environments based on Lidar data under assumption of no uncertainty
\item Extensive simulation experiments of the proposed algorithm 
\end{enumerate}

The algorithm requires a sharing of data amongst the agents and assumes that the robots are equipped with depth sensors (RADAR,LiDAR). An advantage of the proposed algorithm is a continuous time parametrization of the trajectory generation problem with discrete inter-robot collision avoidance. Moreover,the trajectory optimization is solved as a convex optimization problem.

The rest of the paper is organised as: Related works are presented in Section \ref{Related works}. A formal problem definition and assumptions are provided in Section \ref{Problem formulation}.  Section \ref{Convex} details the formulation of the safe region following while the obstacle representation is explained in \ref{local map}. The trajectory optimisation formulation is given in Section \ref{trajectory generation}. The results are discussed in Section \ref{Results} The paper is concluded in Section \ref{Conclusions}.

\begin{figure*}
\centering
\subfloat[][]{\includegraphics[width=0.18\textwidth]{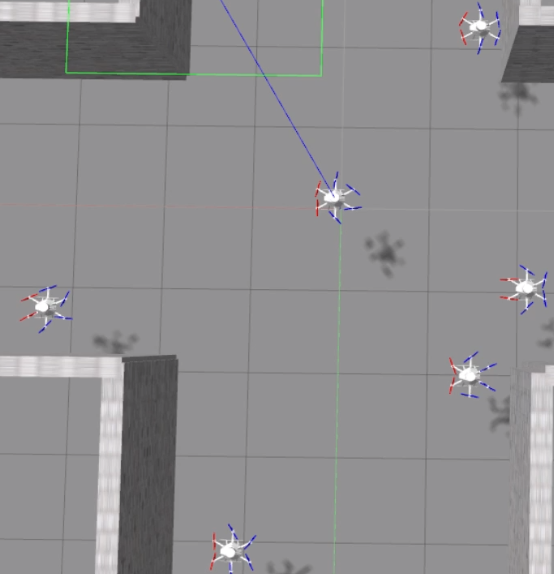}\label{maneuver/maneuver1}} 
\hspace{2mm} \subfloat[][]{\includegraphics[width=0.17\textwidth]{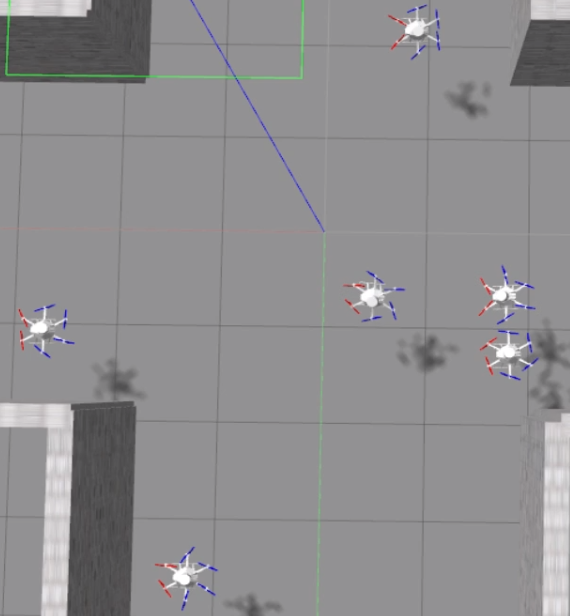}\label{maneuver2}} 
\hspace{2mm} \subfloat[][]{\includegraphics[width=0.195\textwidth]{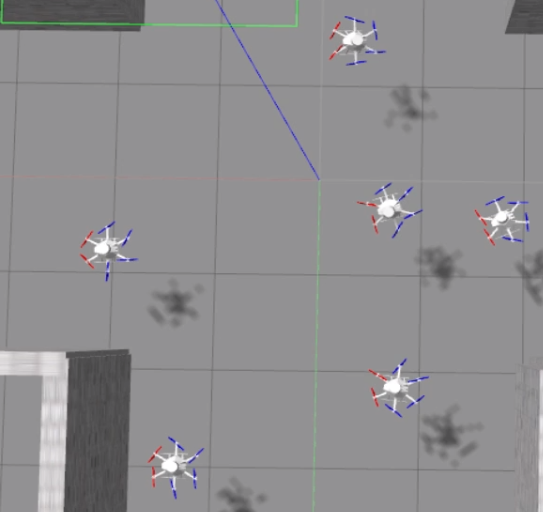}\label{maneuver3}} 
\hspace{2mm} \subfloat[][]{\includegraphics[width=0.17\textwidth]{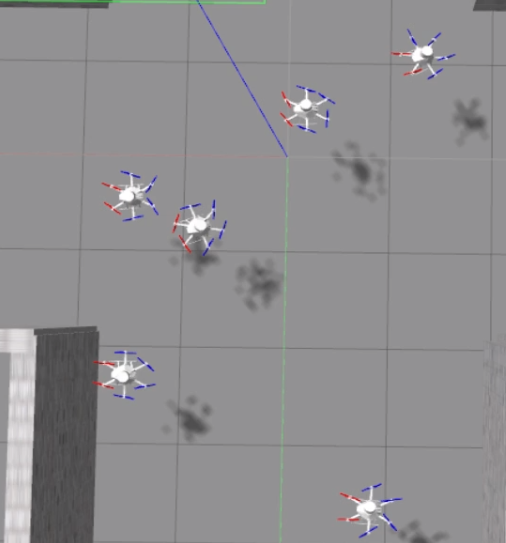}\label{maneuver4}} 
\hspace{2mm} \subfloat[][]{\includegraphics[width=0.18\textwidth]{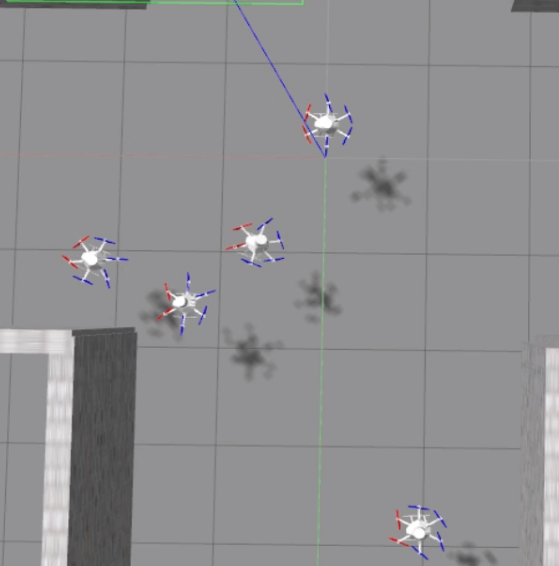}\label{maneuver5}} 
\caption{A sequence of images showing Six fireflys during different stages of the experiment. }
\label{differentposes}
\end{figure*}

\section{Related Work}
\label{Related works}


In \cite{tang2018hold}, a centralized multi robot trajectory planner for obstacle free environments was proposed utilizing tools from non linear optimization and calculus of variation and exploiting a two step process with first step accounting for the collision avoidance.  
A centralized mixed integer programming based approach to multi robot path planning was proposed by \cite{schouweenars2001mixed} with collision avoidance accounted for using binary integer constraints.

In \cite{sutorius2017decentralized}, a decentralized method was proposed using polygonal representations for obstacles utilizing a switched systems to achieve collision-avoidance. Generating regions of free space for the robots was attempted using voronoi cells was tried out by \cite{2017fast}. They utilize a Receding Horizon control based approach which they are able to formulate as a convex quadratic program.

Distributed collision avoidance for multi robot systems have also been attempted \cite{Berg2008} in a method called reciprocal velocity obstacle, exploiting the concept of Velocity obstacles proposed by \cite{Fiorini1998}. This approach assumed other agents continued movement in a straight line with collisions accounted in future by relative velocities.\cite{mora2018cooperative} proposed a collaborative collision avoidance for non holonomic robots with grid based mapped environments  whilst respecting the vehicular constraints and also accounting for potential tracking error bounds for the robot that is planning. 
  
A fully distributed algorithm for navigation in unknown environments was proposed by \cite{Zhou2017real} using incremental sequential convex optimisation for trajectory generation in a model predictive control setting. Distributed re-planning for multiple robots with each robots having different planning cycles in known obstacle filled environments was attempted by \cite{Bekris2017Safe}.The transmission of previously generated trajectories and plan trajectories while avoiding collisions with these trajectories and utilize conservative approximations to account for deviations from these expected trajectories by the other robots.

\section{Problem formulation}
\label{Problem formulation}
Consider $N \in \mathbb{N} $ robots in a 2 dimensional workspace with an unknown number of obstacles and their sizes. The position of i\textsuperscript{th} robot is represented by $ P_i \in \mathbb{R}^2 $. Each of the N robots have a set of time stamped desired poses. We also assume every robot gets to know the current time stamped state of other robots in the vicinity at frequent intervals. We take the state as $x_i = [P_i \hspace{2mm} v_i \hspace{2mm} a_i ]$ . Where $ v_i$ is the velocity of i\textsuperscript{th} along the two axes, $ a_i$ is acceleration of i\textsuperscript{th} robot. Therefore, a common reference frame for all the robots is a requisite. Then it is required that the robots go from their current positions through all their desired poses at as close as possible to the specified timestamps as possible. That is for all the robots, a trajectory is to be planned that ensures that robot traverses from its current position to a within a region from the desired position within the specified time while not colliding with any of the other robots and/or obstacles in the environment. 

Furthermore, we assume that each robot is equipped with a rangefinding based sensor that can give the depth information of the obstacles and that the depth sensor only perceives obstacles within a sensing region. We also assume that the robots do not know the number of robots in the environment and their desired poses and just utilize the states received by them for forward prediction.

The constraints on the overall system are:
\begin{enumerate}
\item The current state of all the robots
\item The states remain within the feasible set of the respective robots
\item The positions of any robots at any time from $t_1$ to $t_2$ should not coincide.(Inter-Agent Collision Avoidance)
\item The position of an obstacle and a robot should not overlap (Obstacle Avoidance)
\end{enumerate}

\section{Obstacle Detection}
\label{local map}
A local map representation is implemented for obstacles in the environment. The robot is assumed to be employed with a 2D laser range finder, whose data is utilized to find the shape of the object and its center. The rangefinder data from the lidar provides the distance of the obstacles.

The detection of reflection pattern is achieved by using a search through the data that is available in the lidar, that isn't infinity(no reflection back). The  set of reflections together between a non reflection form a single obstacle. as the resolution of a lidar is known beforehand and hence, using the resolution and distance to the obstacle from the robots current position, the obstacle point's position can be comprehended by :

\begin{equation}
\begin{split}
x_{obs}= l*cos(\theta) + P_{x_{ego}} \\
y_{obs}= l*sin(\theta) + P_{y_{ego}}
\end{split}
\label{distance}
\end{equation}
 
This results in a primitive but a simple structure of the robot's obstacles to be noted at. The selection of circle is motivated by the reason that higher sided convex polygons can be approximated accurately by circles of appropriate radius. Moreover, it is easier to approximate from observing a shape partially.


The robot's obstacles with a threshold formed by their radius are utilized for formulating the obstacle avoidance. That is the obstacle's center and it's radius are used for obstacle from the lidar data.  This method while being primitive, allows for simpler obstacle representations for a robot functioning in a 2 dimensional environment. Hence, the obstacles can be easily stored as with their center and sizes. It also has a drawback of over approximating the obstacles at times by formulating drastic radii.

\section{Convex Region}
\label{Convex}
For the formulation of the safe regions for the robots to plan trajectories within, We forward simulate other robot's trajectories utilizing their current states.

\subsection{Forward Simulation} 
Forward simulation is done utilizing the current state of the robot. The forward simulation of the robots is done by:

\begin{equation}
P_i(t) =  P_i(t_\delta) + v_i(t_\delta)(t-t_\delta) + a_i(t_\delta)(t-t_\delta)^2
\label{prediction}
\end{equation}

where $ t_\delta $ is the time stamp of the robot's transmitted state.

The forward simulation is done for specified time horizon $t_h$  based upon the discretization $\tau$.

\subsection{Accommodating the size of robots}
\label{size}
At each of the discretized time points, utilizing the transmitted size of the robots, we formulate regions depending on the transmitted data. For simplicity that if three numbers are sent, The robot is modeled as a cuboid or a cube, two numbers a cylinder and a single number as a sphere. In the case of three numbers, robots are modeled as a square of diagonal of largest side, thereby allowing the robot to rotate freely. In case of two or one, the robot is modeled as a circle. A robot inflated to its size is represented as $\mathsf{R}_i$ .
\begin{equation}
\mathsf{R}_i = 
\begin{cases}
A_i P_i(t) \leq B & Square \\
\norm{{P}_i(t)-\textbf{P}_i(t)} \leq r & Circle
\end{cases}
\end{equation}

Where $B $ is formulated as $P_i(t) \pm \sqrt[]{2}\max(l)$ and $r$ is the radius of the robot

\begin{figure}
\includegraphics[width=0.5\textwidth]{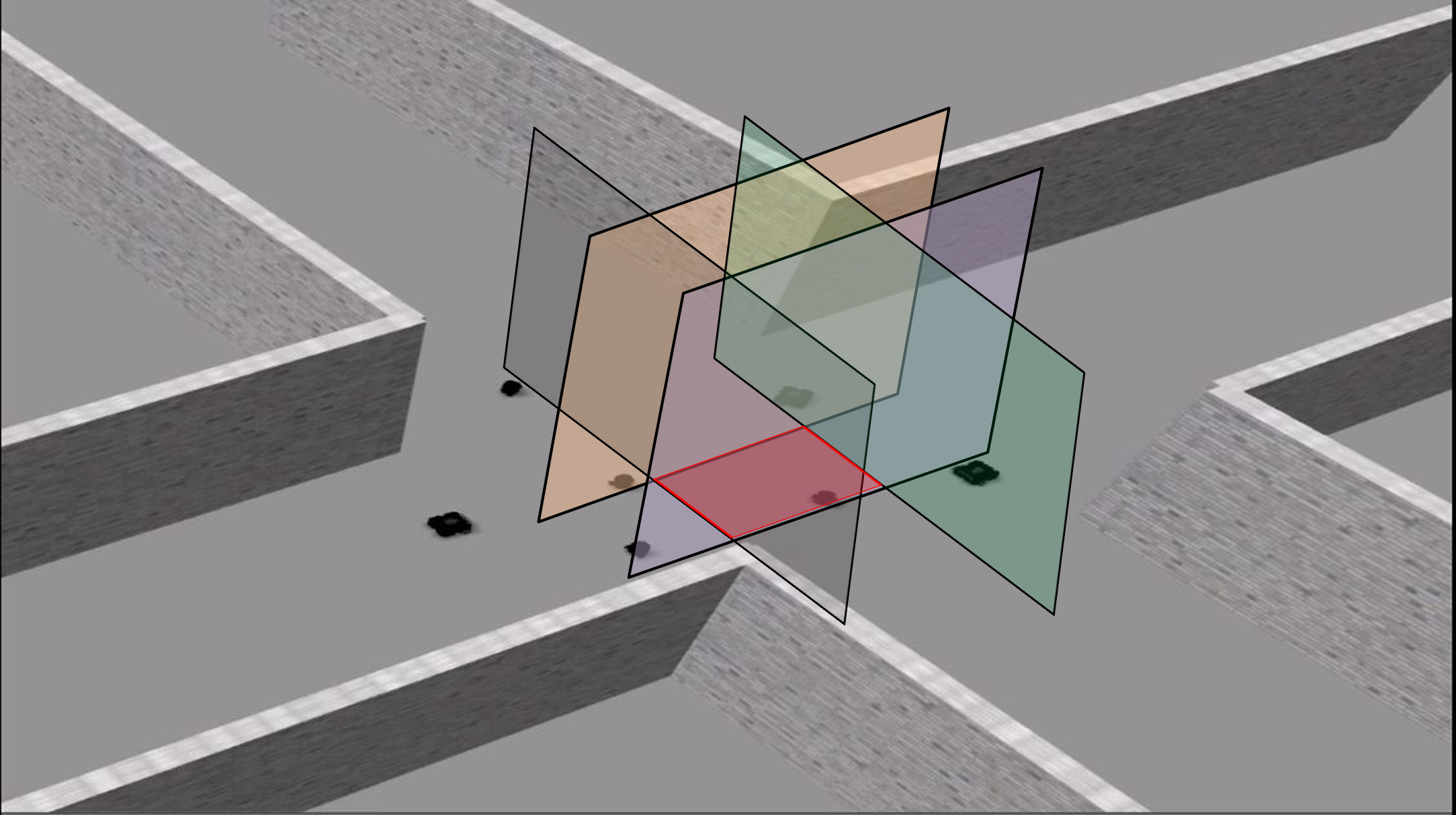}
\label{convex region}
\caption{The red box showcases Collision free safe region for a robot to be constrained within formulated by the positions of the other robots and the transparent planes showing the separating hyperplanes for other robots. }
\end{figure}

\begin{algorithm}
\label{euclid}
\caption{Safe Convex Region}
\begin{algorithmic}
\State \textbf{Given:}
\State \hspace{3.5mm} Agents' current state and size 
\State \hspace{3.5mm} Number of robots (N)
\State \hspace{3.5mm} Moving Volume at specified time points
\For{t=$t_\delta$ to $t_h$}

\For{$i=1$ to N }
\State $P_i(t) =  P_i(t_\delta) + v_i(t_\delta)(t-t_\delta) + a_i(t_\delta)(t-t_\delta)^2$
\State Approximate size of the robots according to \ref{size}
\State Get the intersection of hyperplanes
\EndFor
\EndFor
\end{algorithmic}
\Return Safe Regions
\end{algorithm}

\subsection{Hyperplane}
Considering a time $t_\tau $ between $t$ and $t_h$, supporting hyperplanes are formulated for all $\mathsf{R}_i$ that are within their appropriate moving volumes and as $\mathsf{R}_i$ convex and a supporting hyperplane hence, exists. \cite{boyd2004convex}. A support hyperplane for $\mathsf{R}_i$ can be formulated as   
\begin{equation}
\eta_i(\mathsf{R}_i) \leq \eta_i(\mathfrak{r}_i^0)  
\end{equation}
where $\mathfrak{r}_i^0$ is the boundary of the set $ \mathsf{R}_i $

The intersection of all the support hyperplanes is convex polyhedron as an intersection of convex hyperplane is convex. We Constrain the robot to remain within the generated polyhedron at the discretized time points with polyhedron at each time point represented as $\mathcal{H}(t_{disc}) \leq h $. But as this region constrains the overall robot as a point we subtract the robot's dimensions from the $h$ to constrain the robot to be within the region. the convex regions at specific time points. As the robot's modeled so that it's size is invariant to rotation, dilating the robot according to it's shape of the robot.

\section{Trajectory Generation} 
\label{trajectory generation}
The generation of trajectory by the robot can be formulated as an optimization problem that tries to optimize the system while ensuring that

\begin{argmini!}
  {\textbf{\textit{x}}}{C_{int} + C_{final} + C_{collision} }{}{}
  \addConstraint{\textbf{\textit{x}}(t_1)}{=x_0}{}
  \addConstraint{\sum_{t_{disc}=1}^{\frac{t_h}{\tau}}}{\mathcal{H}(t_{disc}) \leq h}{ ,\hspace{2mm}  t_1 \leq t \leq t_2 }
  \addConstraint{\underline{\ddot{x}}\leq}{x \leq \bar{\ddot{x}}}{}
  \label{cost functional}
\end{argmini!}

While the above mentioned problem is a continuous time problem and thereby a infinite dimensional problem. To transform the problem into a finite dimension, the trajectory is represented by polynomial in each dimension.

\begin{equation}
x(t)= \sum_{j=0}^{2n-1}\alpha_j t^j
\label{decision with degree}
\end{equation}

\begin{equation}
\mathcal{D}_i = [\alpha_0 \hspace{1mm} \alpha_1 \hspace{1mm} \alpha_2 \hspace{1mm} \cdots \alpha_{2n-1}]^T
\label{per_vehicle}
\end{equation}

\subsection{Objective Function}
$C_{int}$ is the integral cost functional that specifies the objective for the derivative over the integral. $C_{final}$ is the cost at the end of the time horizon. $ C_{collision} $ is the collision cost for static obstacles along the trajectory. 

\subsubsection{Derivative Cost}
Derivative penalty utilized is square of the integral of n\textsuperscript{th} derivative and n-1\textsuperscript{th} squared over time horizon t\textsubscript{h}:
It is represented as:
\begin{equation}
C_{int}=\int_{t}^{t_h} Q_{\text{n-1\textsuperscript{th}}}\norm{\frac{d^{n-1}\textbf{\textit{x}}}{dt^{n-1}}}^2 + Q_{\text{n\textsuperscript{th}}} \norm{\frac{d^{n}\textbf{\textit{x}}}{dt^{n}} }^2 \hspace{1mm} dt 
\label{Derivative Cost}
\end{equation} 

Where $Q_{\text{n\textsuperscript{th}}}, Q_{\text{n-1\textsuperscript{th}}}$ are tuning weights for the objective.

As the time horizon is known before hand and the initial time and position are known, We can formulate this cost with a closed form solution as a Quadratic Objective with the decision vector as:
\begin{equation}
D^T H(t+t_h) D
\end{equation}
With $H(t+t_h)$ formulated by integrating Equation \ref{Derivative Cost} and substituting $ t$, $t_h $ and separating according to the coefficients of the polynomial.

\subsubsection{End Cost}
We add an end point quadratic cost for the final position along the trajectory as a soft constraint for two reasons. One, to allow to robot to plan appropriate trajectory if a robot or an obstacle is occupying or blocking the path directly to the end point. Two, in scenarios where the robot's end pose's time stamp is beyond it's trajectory planning horizon, this cost tries to drive the robot as close as possible to it, while ensuring the dynamic limits aren't violated by the hard dynamic constraint

\begin{equation}
C_{end}=(x_{des}-\textbf{\textit{x}}(t_h))^2 Q_{final}
\end{equation}

The final position is penalized but if required addition penalties on velocity, acceleration can be added. The cost can be reformulated with respect to the decision variables resulting in:

\begin{equation}
D^T H(\text{Fin}) D + F(\text{Fin})^T D
\end{equation}

\subsubsection{Collision Cost}
For the generated trajectory to be collision free with respect to the obstacles in the environment,the following penalty is used:

\begin{equation}
C_{collision}=\int_{t}^{t_h} Q_{Obs}c(x(t))v(t)\hspace{1mm} dt 
\label{Obstacle Cost}
\end{equation} 
Where
\begin{equation}
c(x) = \frac{x(t)-x_{obs}}{\exp^{K_p(d(x)-\rho)}d(x)} 
\label{Pieceswise obstacle}
\end{equation}


Here $d(x)$ is euclidean distance to each obstacle. Similar cost functions for collision avoidance have been utilized for collision avoidance for autonomous cars \cite{pmpc}, aerial robots \cite{rob}.$ K_p$ is a smoothness tuning parameter that allows to increase or decrease the smoothness of the collision cost. 

For efficient optimization, a quadratic approximation of the obstacle cost around the previous optimized trajectory is used resulting in:

\begin{equation}
D^T H(\text{Obs}) D + F(\text{Obs})^T D
\end{equation}

\subsection{Constraints}
The trajectory optimisation is constrained by the derivatives of the trajectory staying within the feasible limits(Dynamic Constraints), Staying within the convex region and way-points.

\subsubsection{Waypoint constraints}
The trajectory has to also pass through the given time stamped poses along the trajectory, This results in linear equality constraints on the trajectory.

\begin{equation}
A_{way}D=P
\end{equation}
where $P$ is the stack of poses at their time.

Moreover, as the end pose's time is given in this scenario, the polynomial while minimizing the costs only for the specific time horizon also ensures that the robot reaches the end goal at the desired time stamp 

\subsubsection{Convex Region}
We require that the generated trajectory also remains within the feasible convex region generated at the specific time samples. This constrain is formulated as 
\begin{equation}
\sum_{t_{disc}=1}^{\frac{t_h}{\tau}}\mathcal{H}_i(t_{disc}) \mathcal{T}_i D \leq h
\end{equation}

where $\mathcal{T}$ is the map from the polynomial coefficients to the positions. As both $\mathcal{H} \& \mathcal{T} $ are linear with the polynomial coefficients this results in a convex constraint.

\subsubsection{Dynamic constraints}
Dynamic constraint on the robot is an infinite dimensional and hence we apply the constrains at specific points on the trajectory. This results in added Inequality constrain on the system

\begin{equation}
\sum_{i=1}^n \underline{d} \leq  A_{dyn}D \leq \bar{d}
\end{equation}

Where $n$ is the number of discrete points wherein the constraints are added and $\underline{d} \& \bar{d}$ represent the minimal and maximal limits of the derivatives.

\begin{figure}
\includegraphics[width=0.5\textwidth]{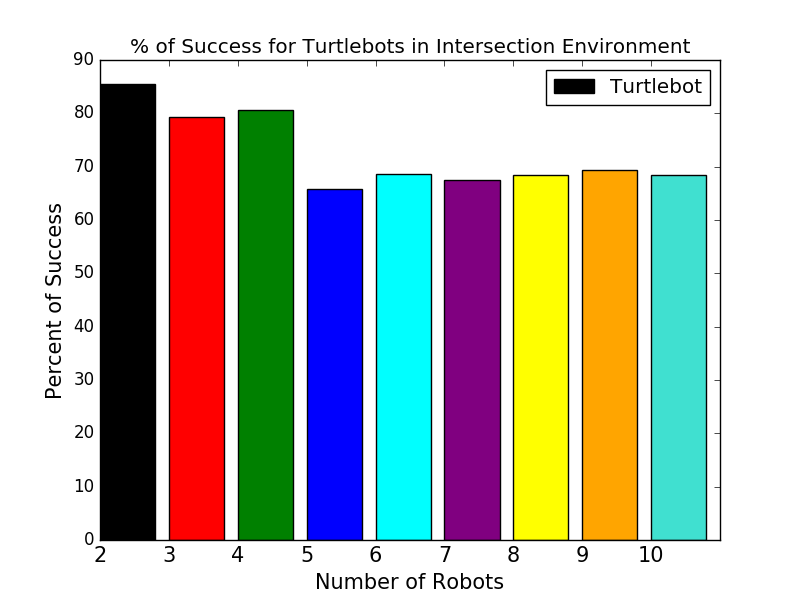}
\caption{The success rates of the algorithm for Turtlebot in the intersection like environment}
\label{graph10}
\end{figure}

The resulting optimization problem can be formulated as a Non Linear Program

\begin{mini}
  {D}{D^T H_{net} D + F_{net}^T D}{}{} 
  \addConstraint{A_{Cont}D}{=0}{}
  \addConstraint{A_{way}D}{=P}{}
  \addConstraint{\sum_{t_{disc}=1}^{\frac{t_h}{\tau}}\mathcal{H}(t_{disc}) \mathcal{T} }{D \leq h}{}
  \addConstraint{\sum_{i=1}^n \underline{d} \leq  A_{dyn}D}{\leq \bar{d}}{}
  \label{trajectory NLP}
\end{mini}

Where $H_{net}$ is formulated by $H(\text{Obs}) + H(\text{Fin})+H(t+t_h)$ and $F_{net}$ by $F(\text{Fin})+ F(\text{Obs})$

The Non Linear Program in Equation \ref{trajectory NLP} is a Convex QP and can be solved using available solvers

\section{Results and Discussion}
\label{Results}
The algorithm was implemented in C++ and integrated into Robot Operating System(ROS) and tested on a workstation with Intel Xeon E5 1630v5  processor, 32GB of RAM and a Nvidia Quadro M4000 GPU. A degree of $2n$-1 was utilized for the polynomials with $t_{h}$ being three seconds. We utilized a $\tau$ of 0.1 seconds for the other agents' prediction. The threshold distance for the obstacle is kept at 0.75m.The algorithm was run at frequency of 25Hz. For solving the QP, qpOASES\cite{qpoases} was used.  

To handle infeasible QPs that arise due to the inequality constraints,  We utilize a two step process for the same. In the initial step, we apply the previous solution for some time and in second step relax the dynamic constraints. 

The proposed algorithm was tested with two sets of robots one a Turtlebot 3 Burger and Waffle (utilizing the native sensor suite available on these robots) and two, AscTec  Firefly and Neo using the a high fidelity simulator \cite{rotors} with robots mounted with Velodyne Puck. The weight of the Velodyne Puck was modified to allow firefly to fly with it. Moreover, Velodyne Puck is a 3D sensor but we limited the sensing to a 2D region. The experiments included both homogeneous and heterogeneous interactions in an intersection like environment, a cross road-like structure which was formed with the help of walls. We utilize an intersection-like environment as it is an important usage of the labeled multi robot problem. 

\begin{figure}
\includegraphics[width=0.5\textwidth]{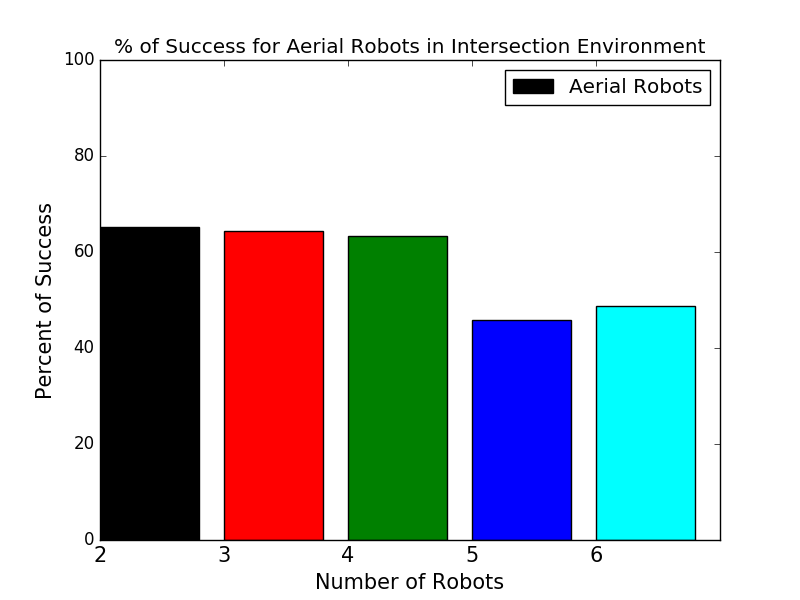}
\caption{The Success rates of the algorithm for Aerial Robots in the intersection like environment}
\label{graph6}
\end{figure}

\begin{figure*}
\subfloat[][Executed trajectories for Six fireflys at the intersection-like environment ]{\includegraphics[width=0.5\textwidth]{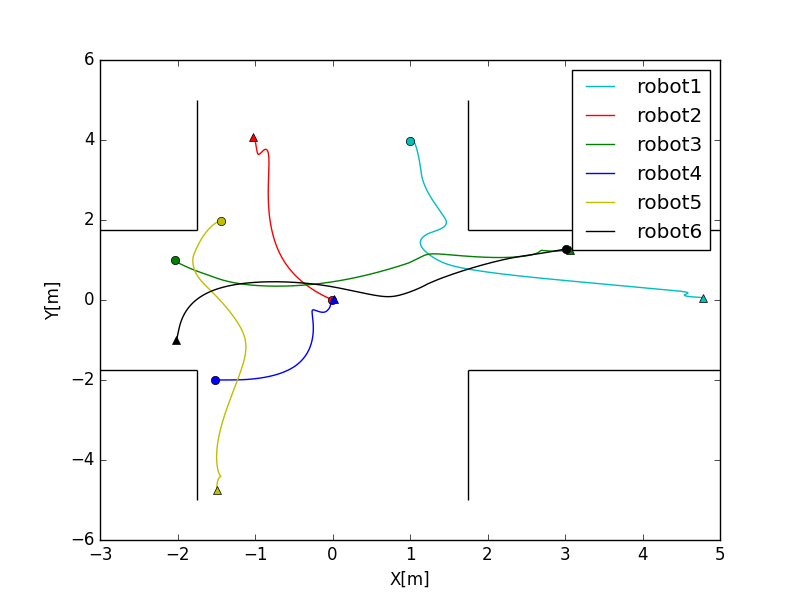}\label{lidar2}} 
\subfloat[][The executed trajectories for eight aerial robots(Six fireflys and two Neos) in the intersection-like Environment]{\includegraphics[width=0.5\textwidth]{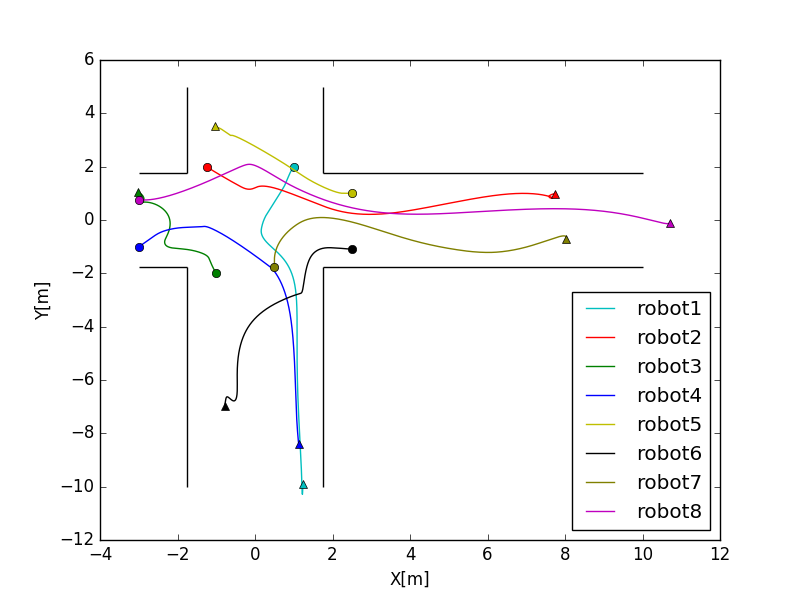}} 
\caption{Trajectories for aerial robots through the intersection like environment. The solid black lines are the walls of the intersection-like area}
\label{traj}
\end{figure*}

The robots were spawned randomly. The desired timestamped poses are predefined in for every robot. For the multirotors, the altitude was fixed at 1. For the ground robots, the generated trajectory is tracked using the MPC proposed in \cite{mpc} which was solved using qpOASES \cite{qpoases} with a time horizon of 2 secs and a discretization of 10Hz. The aerial robots tracked trajectories with the controller proposed in \cite{taecontroller}. The failure rates are shown in Fig \ref{graph10} and Fig  \ref{graph6}. The trajectories of Aerial robots at the intersection-like environment is shown in Fig \ref{traj}.

\subsection{Discussions} 

During the course of simulation, in a few experiments, the robots collided with the obstacles. The collisions in some cases are due to the inaccurate representation of the obstacles and also the unaccounted sensors noise. Moreover, the LIDAR measurements are available at 5Hz whereas the trajectory optimization algorithm runs at 25Hz. The usage of a single polynomial is also a potential cause for this problem. The algorithm also shows some unnecessary non smoothness in its transition towards the final end pose. The safe region generation is conservative but also results in available free-space being neglected. This at times results in the QP being infeasible.

\section{Conclusion}
\label{Conclusions}
A decentralized algorithm for collision free navigation of multiple robots in unknown two dimensional environments was proposed in this work. The proposed algorithm parametrized trajectories by a time parameterized polynomial and generated safe regions based on the prediction of other robots in the environment. A method for obstacle representation was utilized that allowed for simpler collision avoidance with the obstacles. The proposed method was tested extensively in simulations using gazebo for up to eight aerial robots and ten turtlebots.

Utilizing piecewise spline representations of non uniform B-Splines or bezier curves are another avenue for research. The collision representation is a discrete time representation and a continuous time representation of the collision is an avenue for future research. Moreover, sophisticated models of prediction can be utilized for a better prediction accuracy. The sensors used had low update rates. Utilizing faster sensors or utilizing RGB-D or Stereo cameras for local maps is an another important possible extension. Moreover, incorporating a higher variety of primitives for the obstacles will allow for a much more accurate obstacle representation. Another avenue for future research would be test the capabilities of the algorithm in an unstructured environment.

\section*{Supplementary Material}
The experiments in Gazebo can be found at \href{https://youtu.be/wGu0GMOTeH8}{https://youtu.be/wGu0GMOTeH8} \\and \href{https://www.youtube.com/watch?v=JRrxJCXMD_I}{https://www.youtube.com/watch?v=JRrxJCXMD\_I}

\addtolength{\textheight}{-2cm}   



\bibliographystyle{IEEEtran}
\bibliography{references,refer}

\end{document}